\documentclass[conference,letterpaper]{IEEEtran}
\IEEEoverridecommandlockouts
\usepackage{cite}
\usepackage{amsmath,amssymb,amsfonts}
\usepackage{graphicx}
\usepackage{textcomp}
\usepackage{xcolor}

\usepackage{booktabs}
\usepackage{makecell}
\usepackage{array}
\usepackage{tablefootnote} % for table footnotes
\usepackage{subfigure}
\usepackage{adjustbox}
\usepackage{stfloats}
\usepackage{hyperref}
\usepackage{algorithm}
\usepackage[noend]{algpseudocode} % 取代 algorithmic
\usepackage[utf8]{inputenc}
\usepackage{lipsum}
\usepackage{tikz}
\usepackage{pgfplots}
\pgfplotsset{compat=1.18}

\usepackage[letterpaper, left=0.755in, right=0.755in, top=0.755in, bottom=0.755in]{geometry}

\usepackage{titlesec}
\titlespacing*{\section}{0pt}{6pt plus 1pt minus 1pt}{2pt plus 1pt}
\titlespacing*{\subsection}{0pt}{4pt plus 1pt minus 1pt}{1pt plus 1pt}
\titlespacing*{\paragraph}{0pt}{2pt}{0.5em}

\def\BibTeX{{\rm B\kern-.05em{\sc i\kern-.025em b}\kern-.08em
    T\kern-.1667em\lower.7ex\hbox{E}\kern-.125emX}}
\begin{document}

\title{\vspace*{0.25in} \textcolor[HTML]{ba55d3}{MORPH-U}: \textcolor[HTML]{ba55d3}{M}ulti-\textcolor[HTML]{ba55d3}{O}bjective \textcolor[HTML]{ba55d3}{R}esilient Motion \textcolor[HTML]{ba55d3}{P}lanning for V2X-Enabled Autonomous Driving in \textcolor[HTML]{ba55d3}{H}igh-\textcolor[HTML]{ba55d3}{U}ncertainty Environments via Simulation
% \\
% {\footnotesize \textsuperscript{*}Note: Sub-titles are not captured for https://ieeexplore.ieee.org  and
}

\author{\IEEEauthorblockN{Shih-Yu Lai}
\IEEEauthorblockA{
\textit{National Taiwan University}\\
Taipei, Taiwan \\
akinesia112@gmail.com}
% \and
}

\maketitle
\vspace{-8pt}

\begin{abstract}
V2X can warn an autonomous vehicle about hazards beyond line-of-sight, but it also brings uncertainty: messages may be delayed, dropped, or even forged. Meanwhile, map knowledge may change during a trip, forcing the vehicle to replan under tight real-time budgets. This paper studies how to make motion planning and low-level control robust to such uncertain, event-driven updates. We present \textbf{MORPH-U}, a CARLA-based closed-loop stack that fuses LiDAR/radar/camera with V2X (CAM/DENM) into a Local Dynamic Map (LDM) and triggers Hybrid-A* replanning when validated hazards or map changes affect the planned route. We expose the planning/control trade-offs via a multi-objective formulation over tracking error, safety margin (minimum TTC), responsiveness, and smoothness, and select operating points using Pareto-frontier analysis. To avoid unsafe replanning from faulty V2X triggers, MORPH-U adds a lightweight Byzantine-inspired acceptance gate that combines a quorum rule with an on-board sensor veto. Experiments in dynamic CARLA scenarios show that V2X-augmented LDM improves downstream safety, Pareto tuning provides controllable accuracy--comfort trade-offs, and the gate prevents replanning under saturated false-DENM injection ($p_{\text{attack}}{=}1.0$).
\end{abstract}


\section{Introduction}
V2X promises earlier hazard awareness than on-board sensing alone, but in practice V2X inputs are uncertain: messages can arrive late, be lost, or be incorrect under realistic wireless conditions and dynamic traffic~\cite{zhang2024v2x,delooz2023adaptive,ribouh2024seecad}. For a vehicle operating near safety margins, such uncertainty directly affects \emph{when} to replan and \emph{how} aggressively to control. In addition, road knowledge can change during execution (e.g., closures or topology updates), which may invalidate the current route and require replanning within a fixed control cycle~\cite{kaljavesi2024carla}.

While prior work studies V2X cooperation~\cite{zhang2024v2x,delooz2023adaptive}, uncertainty-aware perception~\cite{van2023evidential}, and simulation toolchains~\cite{CARLA,justo2024simbusters}, fewer studies connect them \emph{in a closed loop} with measurable planning/control trade-offs under event-driven V2X/map updates, where safety, tracking, and smoothness are inherently competing.
We present \textbf{MORPH-U}, a CARLA-based vehicle-side pipeline that (i) fuses LiDAR/radar/camera with V2X CAM/DENM into a Local Dynamic Map (LDM), (ii) runs Hybrid-A* planning with event-driven replanning, and (iii) executes trajectories using Pure Pursuit and PID control~\cite{CARLA}. On top of this loop, we add two mechanisms that target the above uncertainties: a multi-objective Pareto analysis for selecting operating points across tracking, safety, responsiveness, and smoothness~\cite{deb2002nsga,zitzler2003performance}; and a lightweight Byzantine-inspired acceptance gate that filters V2X-triggered replanning events before they affect the vehicle~\cite{lamport1982byzantine,blanchard2017machine}.

Our contributions are:
\begin{enumerate}
  \item A V2X-augmented LDM that integrates CAM/DENM with on-board sensing and supports closed-loop planning under uncertain external inputs~\cite{zhang2024v2x,delooz2023adaptive,van2023evidential}.
  \item An event-driven Hybrid-A* replanning loop in CARLA that reacts to validated hazards and map changes during execution~\cite{CARLA,kaljavesi2024carla}.
  \item A multi-objective formulation and Pareto-frontier analysis that makes safety--tracking--smoothness trade-offs explicit and selectable~\cite{deb2002nsga,zitzler2003performance}.
  \item A Byzantine-inspired acceptance gate for V2X-triggered replanning, evaluated under injected false-event attacks~\cite{lamport1982byzantine,blanchard2017machine}.
\end{enumerate}

We position MORPH-U as an \emph{integration} contribution: Hybrid-A*, Pure Pursuit, PID, and quorum-style filtering are individually standard. The novelty lies in (i) routing CAM/DENM through a probabilistically-fused LDM that is the \emph{sole} planner interface, enabling clean ablations; (ii) treating planner re-entry as an explicit, gated event rather than continuous re-optimization; and (iii) reporting V2X's downstream effect on a \emph{measured} Pareto frontier rather than at hand-tuned points.

\begin{figure}[t]
\centering
\subfigure[Multi-vehicle urban intersection (S2/S4).]{%
  \includegraphics[width=0.48\columnwidth]{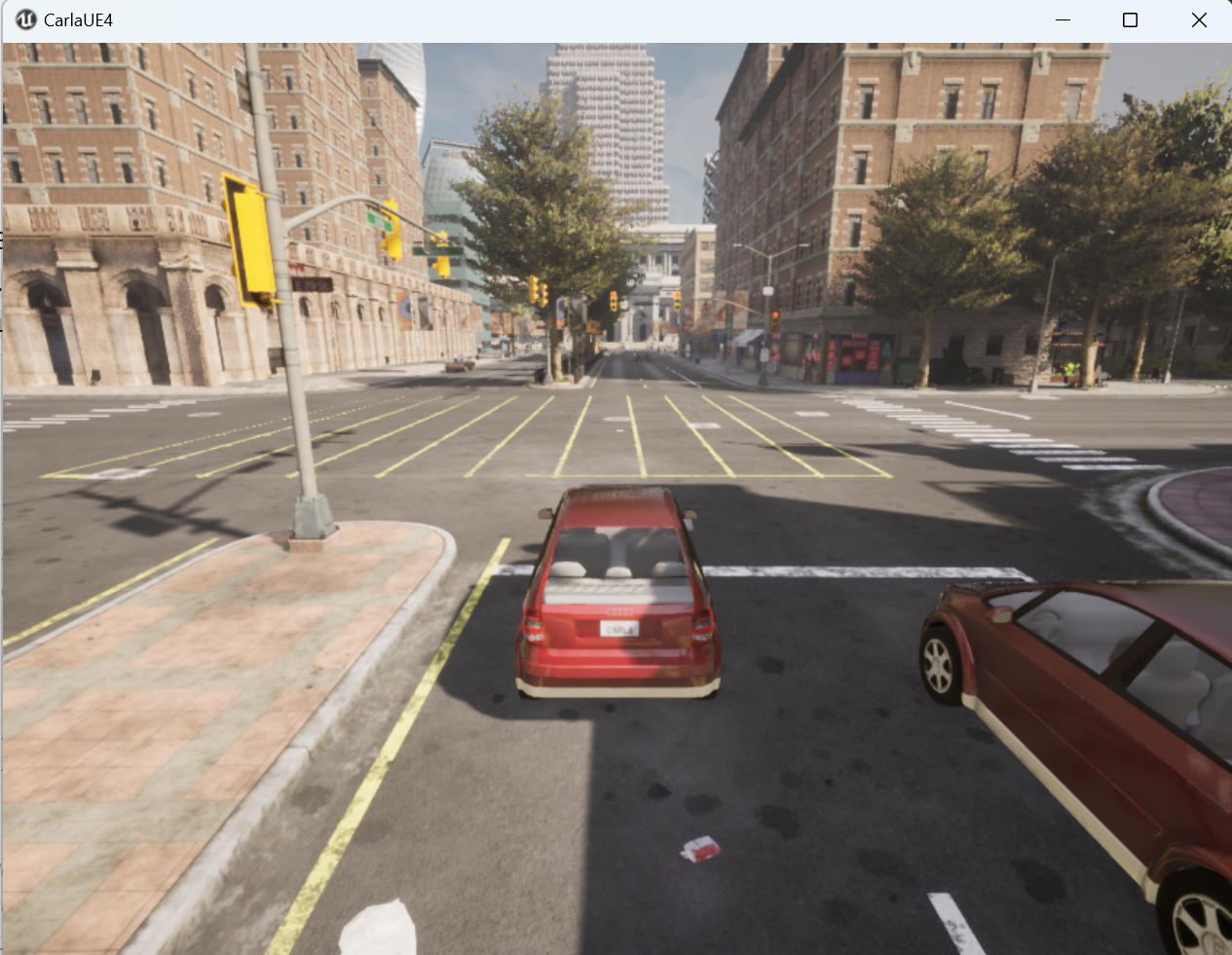}%
}\hfill
\subfigure[Single-vehicle route-following (S1/S3).]{%
  \includegraphics[width=0.48\columnwidth]{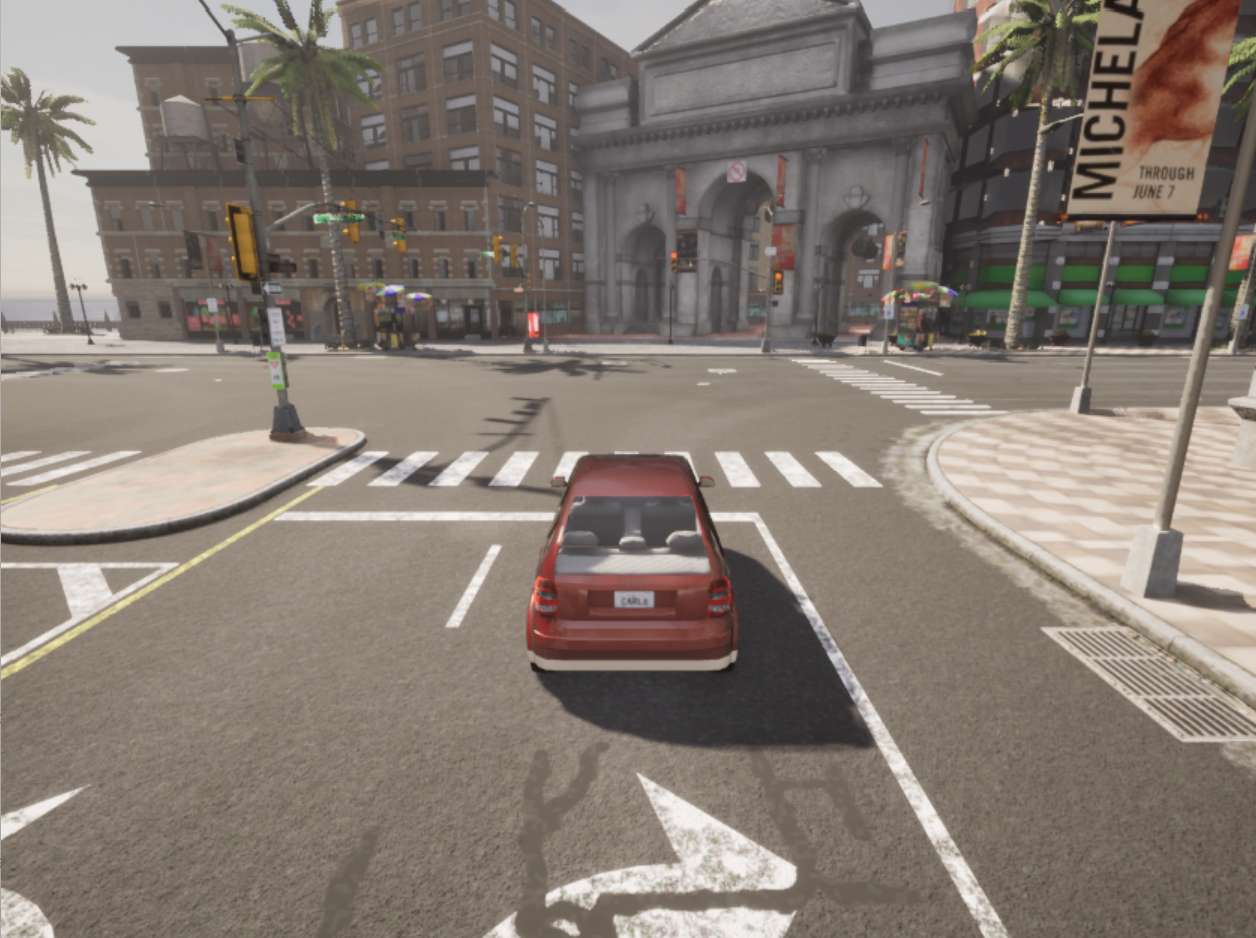}%
}
\caption{CARLA simulator evaluation setups in MORPH-U. \textbf{(a)} Multi-vehicle urban intersection used for V2X-enabled hazard response and faulty-trigger robustness (S2/S4). \textbf{(b)} Single-vehicle route-following used for baseline tracking and Pareto operating-point selection (S1), and as the backbone setting for update-induced rerouting (S3).}
\label{Fig_1}
\end{figure}

We evaluate MORPH-U in CARLA urban intersections under both multi-vehicle traffic and single-vehicle route-following setups (Fig.~\ref{Fig_1}), covering baseline tracking (S1), V2X hazard response (S2), update-induced rerouting (S3), and faulty-trigger attacks (S4).

\section{Related Work}
\label{sec:related}

\subsection{V2X-augmented perception under uncertainty}
V2X has been widely studied as a mechanism to extend perception beyond line-of-sight through cooperative awareness and event-driven safety messages, enabling cooperative and infrastructure-aided perception in dynamic traffic~\cite{zhang2024v2x,delooz2023adaptive,ribouh2024seecad}. 
In parallel, uncertainty-aware fusion has long relied on Bayesian and evidential formulations to represent ambiguity and combine heterogeneous sensing sources~\cite{Bayesian,Dempster-Shafer,mentasti2024tracking,van2023evidential}. 
Learning-based modules further enhance perception and downstream tasks in such pipelines~\cite{deeplabv3plus2018,mentasti2024tracking}. 
These works establish that V2X can improve the world model, but they typically stop short of quantifying how such improvements propagate through a closed-loop planner and controller under event-driven updates.

\subsection{Planning, control, and multi-objective operating points}
Search-based planning and classical controllers remain common baselines for closed-loop autonomy, especially when reproducibility and auditable failure modes are required in safety studies~\cite{CARLA,mu2024pix2planning}. 
However, in realistic settings, safety, tracking, responsiveness, and smoothness are competing objectives that cannot be optimized simultaneously. 
Multi-objective optimization offers a principled lens to expose and select operating points via Pareto-optimal sets and quality indicators~\cite{deb2002nsga,zitzler2003performance}. 
Yet, much of the autonomy literature reports results at a small number of hand-tuned configurations, leaving the trade-off structure implicit and making it difficult to compare planning/control behaviors across scenarios with different uncertainty profiles.

\subsection{Resilience to faulty V2X triggers and map inconsistency}
V2X inputs introduce a new failure surface: incorrect or adversarial reports can trigger unsafe braking or replanning, motivating resilience mechanisms grounded in Byzantine reasoning and robust aggregation~\cite{lamport1982byzantine,blanchard2017machine,chen2017blockchain}. 
Separately, maintaining map consistency for downstream autonomy has been studied through HD map representations and validation/update pipelines, including standards such as OpenDRIVE and related change-handling work~\cite{OpenDRIVE,hdmaps_open_drive2022,Lanelet2,hdmap_verification_no_localization2023,evaluation_hdmap_selflocalization2023,terminology_map_deviations2023,lanemapnet2023,high_integrity_lane_occupancy2023,hdmap_from_noisy_data2023,e_mlp_online_hdmap2023,smartmot2023,kaljavesi2024carla}. 
Simulation and toolchains enable controlled evaluations of such effects in integrated pipelines, including V2X-in-the-loop setups and CARLA-based benchmarking~\cite{CARLA,geller2024carlos,justo2024simbusters,V2X-ROS,Grimm2024CARLA-V2X-Sensor,kaljavesi2024carla,grimm2024contextualfusion}. 
However, fewer studies combine (i) V2X-augmented fusion, (ii) event-driven replanning induced by hazards and map changes, (iii) explicit Pareto trade-offs in planning/control, and (iv) resilience gating against faulty triggers \emph{within the same closed-loop stack}. MORPH-U targets this gap by integrating these components and reporting measurable trade-offs and robustness envelopes across scenarios.

\section{System and Method}
\label{sec:system-method}

\begin{figure}[t]
    \centering
    \includegraphics[width=0.5\textwidth]{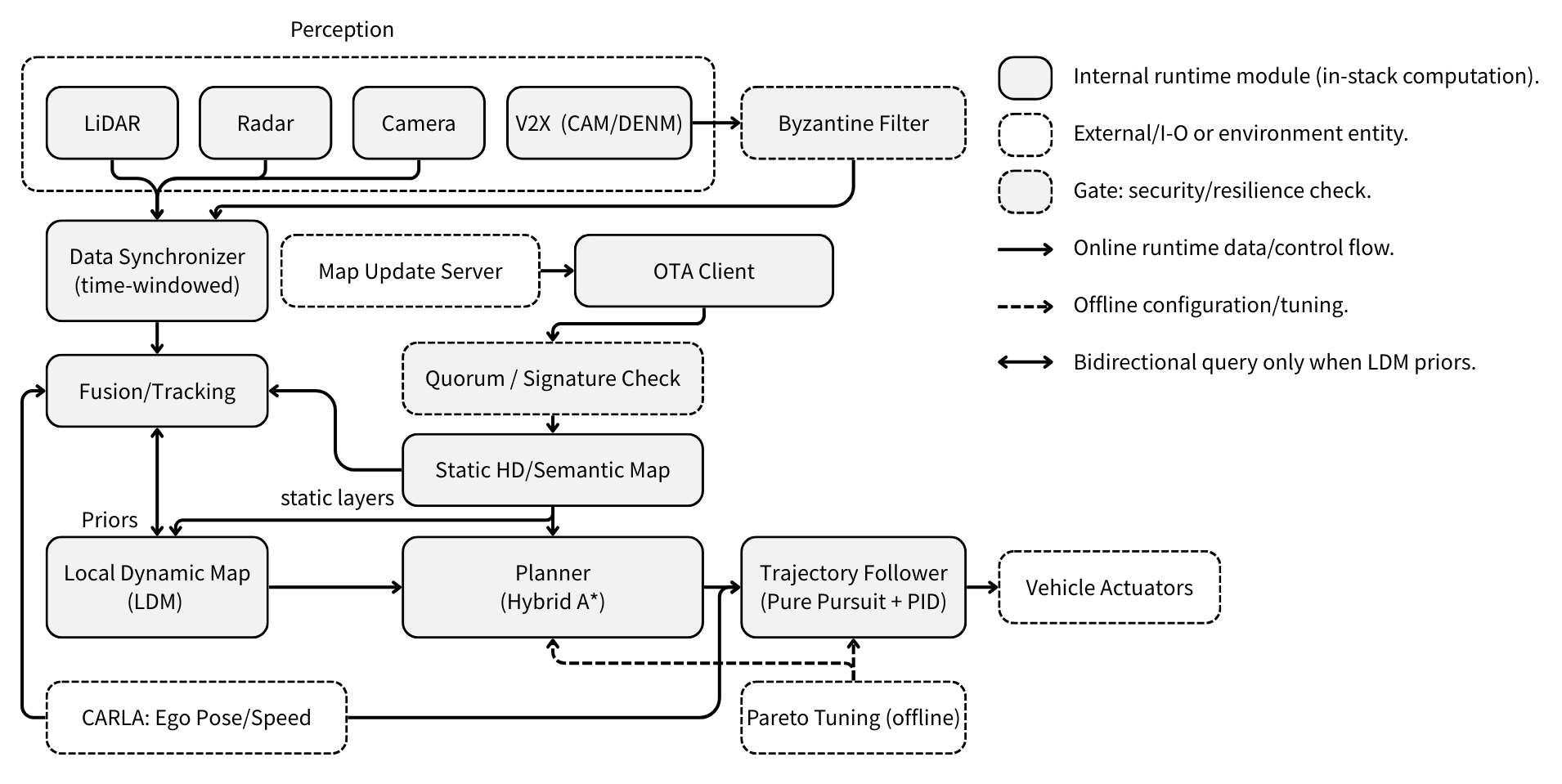}
    \caption{Closed-loop architecture of MORPH-U. Time-windowed synchronization and fusion populate an ego-centric LDM from on-board sensing and V2X (CAM/DENM). Hybrid-A* replans when validated events or knowledge changes trigger replanning. Pareto tuning is performed offline to select planning/control operating points. The acceptance gates (V2X and update path) prevent faulty triggers from reaching the planner.}
    \label{Fig_2}
\end{figure}

\textbf{MORPH-U} is a CARLA-based, vehicle-side closed-loop stack that integrates V2X with multi-sensor perception, search-based planning, and classical control. At each tick, the system (i) buffers on-board sensor frames and V2X messages, (ii) fuses them into a Local Dynamic Map (LDM), (iii) triggers event-driven replanning when hazards or knowledge changes affect the planned route, and (iv) executes the trajectory with a trajectory follower. The design explicitly targets two questions: \emph{(a)} how V2X improves the fused world model used by the planner, and \emph{(b)} how to select stable operating points that balance safety, tracking, responsiveness, and smoothness under uncertainty. Figure~\ref{Fig_2} summarizes the closed-loop execution path and highlights where (i) V2X/map events enter the LDM, (ii) replanning triggers are evaluated, and (iii) the acceptance gate blocks faulty triggers before they affect planning.

\subsection{V2X-Augmented LDM Fusion}
\label{subsec:ldm-fusion}
MORPH-U fuses LiDAR/radar/camera detections with V2X messages (CAM/DENM) into an ego-centric LDM that maintains (i) tracked dynamic objects and (ii) discrete hazard/map events relevant to planning. Let $z_k$ denote an incoming sensor detection or a decoded V2X packet with timestamp $t_k$. A synchronizer emits a time-aligned bundle within a sliding window:
\begin{equation}
\mathcal{S}_t = \{(z_k,t_k)\mid t-\tau_{\text{sync}} \le t_k \le t\}.
\end{equation}
A fusion/tracking module associates detections across modalities and updates the LDM state
\begin{equation}
\mathcal{X}_t = \{\mathcal{O}_t,\mathcal{E}_t,\mathcal{M}_t\},
\end{equation}
where $\mathcal{O}_t$ are tracked objects in the ego-local frame, $\mathcal{E}_t$ are event hypotheses (e.g., DENM hazards, map-change notices), and $\mathcal{M}_t$ are the active static map layers used for planning. V2X messages are \emph{not} consumed directly: each $o\in\mathcal{O}_t$ carries an existence belief $b(o)\in[0,1]$ updated by Bayesian combination of on-board detection likelihoods and authenticated CAM/DENM reports weighted by source reputation, with on-board sensing acting as a veto when $\mathcal{L}_{\text{sensor}}$ contradicts a V2X claim within the same spatio-temporal cell. Event hypotheses $\mathcal{E}_t$ are gated separately by Sec.~\ref{subsec:res-gate} before they can trigger replanning. This LDM is the sole interface consumed by the planner and controller, enabling controlled ablations (sensors-only vs.\ sensors+V2X) and deterministic replay.

\subsection{Planning and Event-Driven Replanning}
\label{subsec:planning-triggers}
Given $\mathcal{X}_t$ and a goal pose, MORPH-U computes a curvature-feasible trajectory using Hybrid-A* over $SE(2)$:
\begin{equation}
|\kappa(s)| \le \kappa_{\max},\qquad \kappa(s)=\frac{\tan\delta(s)}{L},
\end{equation}
with wheelbase $L$ and steering angle $\delta(s)$. Replanning is invoked by three auditable, scenario-controllable triggers: (i) \emph{hazard-on-route}, a validated DENM within a look-ahead horizon on the planned route; (ii) \emph{risk threshold}, a TTC-based predicted-risk excess on the current plan prefix; and (iii) \emph{knowledge change}, an active-map version change after update activation. All triggers route through the gate of Sec.~\ref{subsec:res-gate}, ensuring that V2X/map signals affect the vehicle only via explicit, logged decisions.

\subsection{Multi-Objective Formulation and Pareto Operating Points}
\label{subsec:multiobj}
We expose tunable design variables
\begin{equation}
\boldsymbol{\theta}=\{\text{LA},K_p,K_i,K_d,\ \text{replan thresholds},\ \text{update interval},\ldots\},
\label{eq:theta}
\end{equation}
and evaluate each configuration on an objective vector
\begin{equation}
\mathbf{J}(\boldsymbol{\theta})=
[J_{\text{trk}},J_{\text{sfty}},J_{\text{resp}},J_{\text{smth}},J_{\text{eng}}].
\end{equation}
We instantiate these objectives as:
{\small
\begin{align}
J_{\text{trk}} &:= \mathrm{RMSE}_{\text{lat}}, \quad
J_{\text{sfty}} := \max(0,\ \tau_{\text{sfty}}-\mathrm{TTC}_{\min}), \notag \\
J_{\text{resp}} &:= \alpha t_{\text{V2X}}+\beta t_{\text{upd}}, \quad
J_{\text{smth}} := \mathrm{Var}(\delta)+\gamma\,\mathrm{Var}(\text{thr}), \notag \\
J_{\text{eng}} &:= \text{brake-energy proxy}.
\end{align}
}
We treat collisions as a hard constraint (discard) or a large penalty. For comparability across objectives, we normalize each objective to $[0,1]$ via min--max over the evaluated set: $\tilde{J}_i=\frac{J_i-J_i^{\min}}{J_i^{\max}-J_i^{\min}}$. A configuration $A$ dominates $B$ if $\tilde{\mathbf{J}}(A)\preceq \tilde{\mathbf{J}}(B)$ componentwise and strictly smaller in at least one component; the Pareto set $\mathcal{P}$ is the nondominated subset.

\paragraph{Knee-point selection.}
To select a single operating point for deployment in closed-loop experiments, we choose a knee solution that minimizes distance to the utopia point under zero-collision constraint:
\begin{equation}
\boldsymbol{\theta}^\star \in \arg\min_{\boldsymbol{\theta}\in\mathcal{P}}
\left\lVert \tilde{\mathbf{J}}(\boldsymbol{\theta}) \right\rVert_2
\quad \text{s.t. collisions}=0.
\end{equation}

\paragraph{Hypervolume comparison.}
To compare ablations (e.g., sensors-only vs.\ sensors+V2X; no-update vs.\ update), we report hypervolume $\mathcal{H}$ of the Pareto set with respect to a reference point $\mathbf{r}\succ \max \tilde{\mathbf{J}}$. Larger $\mathcal{H}$ indicates a better achievable trade-off surface. (We report measured frontiers and $\mathcal{H}$ in Sec.~\ref{subsec:pareto-exp}.)

\subsection{Acceptance Gate for Replanning Triggers (Byzantine-Inspired)}
\label{subsec:res-gate}
\textbf{Threat model.} We adopt a \emph{Byzantine-inspired} threat model: V2X reports are authenticated but up to $f$ of $n$ stations may fabricate, replay, or equivocate. We do not claim formal Byzantine fault tolerance; instead we use a quorum-with-sensor-veto rule as a lightweight filter and evaluate it under one saturated injection policy (Sec.~\ref{subsec:scenarios}, S4).

\textbf{Quorum acceptance rule.} For a candidate event $E$ (e.g., DENM hazard or update notice), let $\mathcal{M}_t$ be the set of distinct authenticated reports supporting $E$ within a spatio-temporal window $(R,\tau_{\text{bft}})$, and let $\mathcal{L}_{\text{sensor}}(E)$ be an on-board likelihood (sensor veto). We accept $E$ only if
\begin{equation}
\sum_{m_i\in\mathcal{M}_t} w_i\,\mathbf{1}[m_i \text{ supports }E] \ge \Theta
\quad \wedge \quad
\mathcal{L}_{\text{sensor}}(E)\ge \eta,
\label{eq:quorum}
\end{equation}
where $\Theta$ is chosen to require at least $2f{+}1$ distinct corroborations (e.g., $\Theta=2f{+}1$ with uniform weights), and $\eta$ enforces the sensor veto. Only accepted events trigger replanning. The gate's guarantees are empirical and limited to the evaluated attack class; coordinated, timing-correlated attacks and sensor-veto bypass are out of scope (Sec.~\ref{sec:conclusion}).

\begin{algorithm}[H]
\caption{Pareto Frontier: Nondominated set (fast sweep)}
\label{alg:nondom}
\begin{algorithmic}
\State \textbf{Input:} normalized objective set $\{\tilde{\mathbf{J}}(\boldsymbol{\theta}_k)\}_{k=1}^N$
\State $\mathcal{P}\leftarrow\emptyset$
\For{$k=1$ to $N$}
  \State dominated $\leftarrow$ \textbf{false}
  \For{$p\in\mathcal{P}$}
    \If{$p \preceq \tilde{\mathbf{J}}(\boldsymbol{\theta}_k)$} 
      \State dominated$\leftarrow$\textbf{true}; \textbf{break} 
    \EndIf
    \If{$\tilde{\mathbf{J}}(\boldsymbol{\theta}_k)\preceq p$} 
      \State remove $p$ from $\mathcal{P}$ 
    \EndIf
  \EndFor
  \If{\textbf{not} dominated} 
    \State add $\tilde{\mathbf{J}}(\boldsymbol{\theta}_k)$ to $\mathcal{P}$ 
  \EndIf
\EndFor
\State \textbf{Output:} Pareto set $\mathcal{P}$
\end{algorithmic}
\end{algorithm}

\section{Implementation Details}
\label{sec:implementation}
CARLA runs in synchronous mode with fixed time step $\Delta t$; a synchronizer aggregates timestamped sensor/V2X packets within $\tau_{\text{sync}}$ and emits fused snapshots each tick. Hybrid-A* operates on the LDM and active map layers, while the trajectory follower combines Pure Pursuit (lateral) and PID (longitudinal) with bounded outputs and anti-windup; look-ahead, gains, and clamps are exposed for Pareto tuning. An update client polls a simulated server for map versions; upon activation, static layers refresh and a replanning trigger is issued at the next tick boundary for consistent state across the control cycle.

\begin{table}[h]
  \centering
  \caption{Metrics (units).}
  \label{tab:metrics}
  \resizebox{0.8\columnwidth}{!}{%
  \begin{tabular}{@{}ll@{}}
    \toprule
    Tracking & Lat.\ RMSE (m), Heading (deg), Completion (\%) \\
    Safety & $\mathrm{TTC}_{\min}$ (s), Collisions \\
    Responsiveness & V2X reaction (ms), Update activation (s) \\
    Smoothness & Var(steer), Var(throttle) \\
    LDM fidelity & MOTA, MOTP, ID switches \\
    Resilience (S4) & FPR, FNR, Trigger latency (ms) \\
    \bottomrule
  \end{tabular}}
\end{table}

\begin{figure}[h]
    \centering
    \includegraphics[width=\columnwidth]{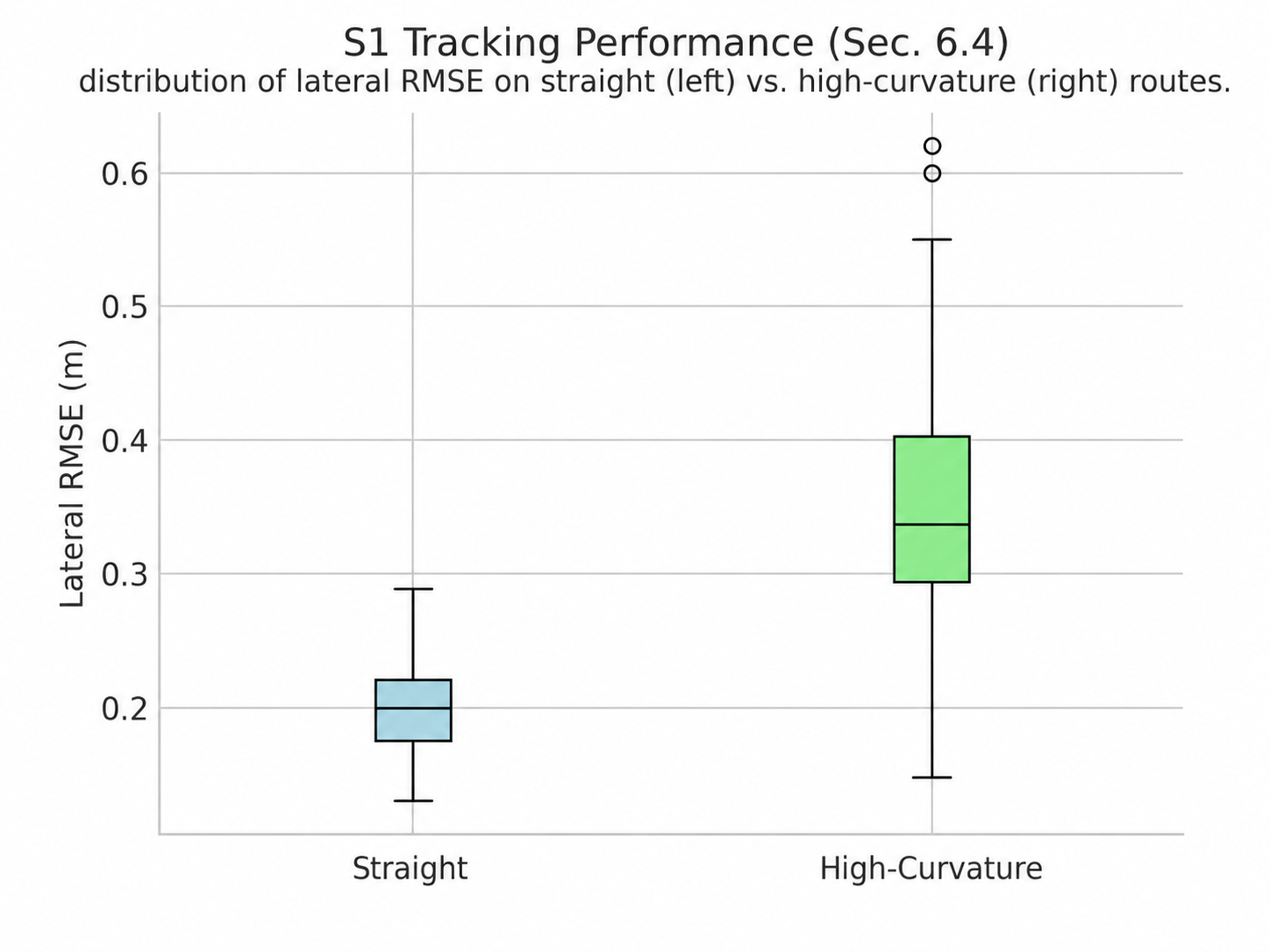}
    \caption{S1 baseline tracking: distribution of lateral RMSE on \emph{straight} (left) vs.\ \emph{high-curvature} (right) routes. The widening tail at high curvature is a known Pure Pursuit artifact; rather than masking it via curvature-dependent speed profiling, we expose look-ahead and gains as Pareto variables (Sec.~\ref{subsec:multiobj}).}
    \label{fig:s1-boxplot}
\end{figure}

\begin{figure}[h]
 \centering
 \begin{tikzpicture}
    \begin{axis}[
      width=0.4\textwidth, height=0.3\textwidth, % Full width
      xlabel={Lateral RMSE (m) $\downarrow$},
      ylabel={Control Smoothness (Variance) $\downarrow$},
      xmin=0.30, xmax=0.60, ymin=0.60, ymax=1.0,
      legend style={draw=none,at={(0.02,0.98)},anchor=north west,font=\scriptsize},
      tick label style={font=\scriptsize},
      label style={font=\scriptsize}
    ]
      % --- "Dominated" 點 
      \addplot+[only marks,mark=o,mark size=1.5pt, gray] coordinates {
         (0.42, 0.81) (0.39, 0.78) (0.41, 0.75) (0.45, 0.77) (0.48, 0.85)
         (0.40, 0.90) (0.50, 0.80) (0.52, 0.78) (0.55, 0.88) (0.43, 0.95)
         (0.38, 0.88) (0.47, 0.82) (0.49, 0.76) (0.51, 0.92) (0.53, 0.81)
         (0.44, 0.79) (0.46, 0.83) (0.48, 0.74) (0.50, 0.89) (0.52, 0.86)
         (0.37, 0.94) (0.43, 0.87) (0.48, 0.91) (0.54, 0.77) (0.57, 0.83)
         (0.41, 0.91) (0.39, 0.83) (0.44, 0.88) (0.50, 0.95) (0.55, 0.92)
      };
      \addlegendentry{Dominated Set}
      
      \addplot+[only marks,mark=*,mark size=2pt, blue] coordinates {
         (0.33, 0.89) (0.35, 0.78) (0.36, 0.72) (0.38, 0.70) 
         (0.40, 0.67) (0.44, 0.64) (0.47, 0.63) (0.53, 0.62)
      };
      \addlegendentry{Pareto Frontier}
      \addplot[mark=star, mark size=4pt, red, thick, draw=none] coordinates {(0.36, 0.72)};
      \addlegendentry{Knee Point (Selected)}
    \end{axis}
 \end{tikzpicture}
 \caption{Pareto frontier (Sec.~\ref{subsec:pareto-exp}): Tracking vs. Smoothness ($N=60$ configurations). We select the 'Knee Point' (red star) for S2-S4 experiments.}
 \label{fig:pareto-frontier}
\end{figure}
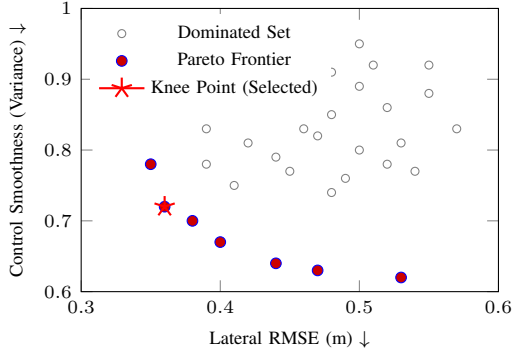

\begin{table}[h]
 \centering
 \caption{Planner Performance.}
 \label{tab:planner-perf}
 \resizebox{\columnwidth}{!}{%
 \begin{tabular}{@{}lccc@{}}
   \toprule
   \textbf{Scenario} & \textbf{Success Rate} (\%) & \textbf{Avg. CPU Time} (ms) & \textbf{Avg. Path Length} (m) \\
   \midrule
   S1 (Baseline) & 100.0 & 28.5 $\pm$ 4.1 & 150.2 \\
   S2 (Hazard) & 96.7 & 35.1 $\pm$ 6.2 & 158.5 \\
   S3 (OTA Reroute) & 96.7 & 38.0 $\pm$ 6.0 & 165.1 \\
   S4 (BFT Attack) & 83.3 & 34.8 $\pm$ 5.9 & 150.2 \\
   \bottomrule
 \end{tabular}
 }
\end{table}

\begin{figure}[h]
    \centering
    \includegraphics[width=0.8\columnwidth]{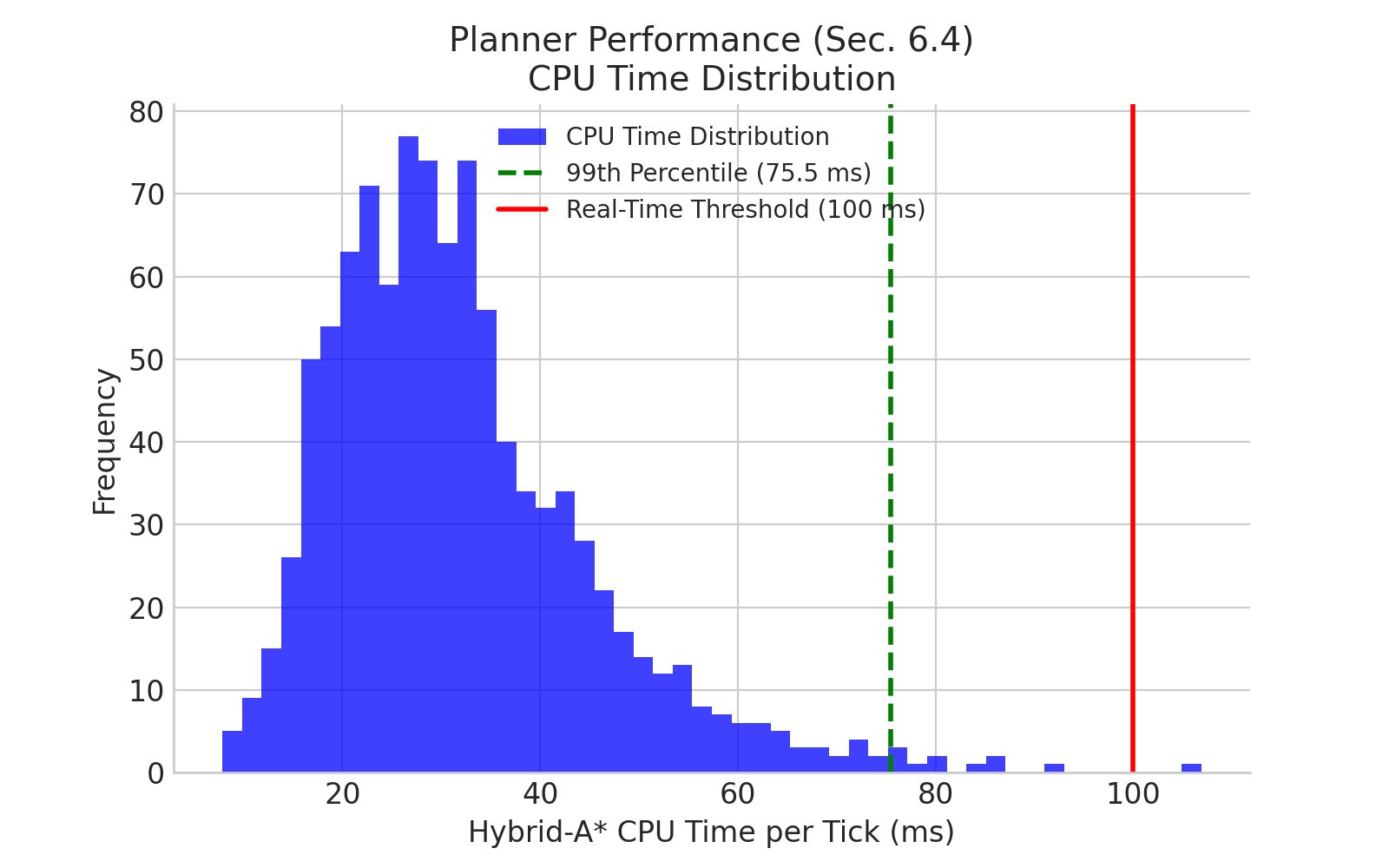}
    \caption{Hybrid-A* runtime distribution per tick. The tail remains within the real-time budget in evaluated scenarios.}
    \label{fig:planner-cpu}
\end{figure}

\begin{table}[h] 
  \centering
  \caption{S2—V2X hazard response and LDM fidelity. Mean $\pm$ SD over 30 seeds. LDM metrics are computed against CARLA ground truth, and mainly reflect CAM augmentation.}
  \label{tab:v2x_results}
  \resizebox{\columnwidth}{!}{%
  \begin{tabular}{@{}p{2.6cm}ccccccc@{}}
    \toprule
    \textbf{Config} 
    & \textbf{Lat.\ RMSE} (m) $\downarrow$ 
    & \textbf{Min TTC} (s) $\uparrow$
    & \textbf{V2X Lat.} (ms) $\downarrow$
    & \textbf{Collisions} $\downarrow$
    & \textbf{MOTA} $\uparrow$
    & \textbf{MOTP} $\uparrow$
    & \textbf{IDSW} $\downarrow$ \\ \midrule
    Sensors only 
    & 0.42 $\pm$ 0.05 
    & 1.30 $\pm$ 0.21 
    & --- 
    & 5/30
    & 0.78 & 0.85 & 22 \\
    + V2X (CAM/DENM) 
    & \textbf{0.35} $\pm$ 0.04 
    & \textbf{1.90} $\pm$ 0.18 
    & \textbf{140} $\pm$ 25 
    & \textbf{0/30}
    & \textbf{0.92} & \textbf{0.94} & \textbf{5} \\
    \bottomrule
  \end{tabular}
  } 
\end{table}

\begin{table}[h]
  \centering
  \caption{S3—OTA-induced reroute. Mean $\pm$ SD over 30 seeds.}
  \label{tab:ota_results}
  \resizebox{0.8\columnwidth}{!}{%
  \begin{tabular}{@{}lccc@{}}
    \toprule
    \textbf{Config} & \textbf{OTA Act.} (s) & \textbf{Replan Time} (ms) & \textbf{Completion} (\%) \\
    \midrule
    No-OTA & — & — & 72.4 $\pm$ 4.7 \\
    OTA enabled & \textbf{1.1} $\pm$ 0.2 & \textbf{38} $\pm$ 6 & \textbf{96.7} $\pm$ 2.1 \\
    \bottomrule
  \end{tabular}
  } 
\end{table}

\begin{table}[h]
 \centering
 \caption{S4—Byzantine Attack Response ($n{=}10$, $f{=}3$, $p_{\text{attack}}{=}1.0$, $N{=}30$ seeds). FNR is w.r.t.\ the injected true hazard $E^\star$.}
 \label{tab:bft_results}
 \resizebox{0.8\columnwidth}{!}{%
 \begin{tabular}{@{}lcccc@{}}
   \toprule
   \textbf{Config} & \textbf{FPR} $\downarrow$ & \textbf{FNR} $\downarrow$ & \textbf{Collisions} & \textbf{Completion} (\%) \\ \midrule
   Sensors only (Baseline) & N/A & N/A & 5/30 & 83.3 $\pm$ 5.1 \\
   V2X (No Filter) & 1.00 (30/30) & 0.00 (0/30) & 0/30 & 0.0 $\pm$ 0.0 \\
   \textbf{V2X + Quorum Filter} & \textbf{0.00 (0/30)} & \textbf{0.00 (0/30)} & \textbf{0/30} & \textbf{96.7 $\pm$ 2.1} \\
   \bottomrule
 \end{tabular}
 } 
\end{table}

\section{Experimental Design}
\label{sec:exp-design}

\subsection{Scenarios}
\label{subsec:scenarios}
We evaluate four scenarios that isolate V2X benefit, knowledge-change replanning, and resilience:
\begin{itemize}
  \item \textbf{S1 (Tracking):} route following with no V2X and no updates; we vary curvature and speed limits to stress tracking and smoothness.
  \item \textbf{S2 (V2X hazard response):} during motion, we inject a DENM hazard (stationary vehicle / road closure) ahead of the ego vehicle; we compare reactions with and without V2X.
  \item \textbf{S3 (Update-induced reroute):} mid-route, a new map version changes topology (add/remove a segment); the vehicle activates the update and replans to complete the route.
  \item \textbf{S4 (Faulty V2X triggers with a true hazard):} based on S2, we inject a \emph{ground-truth} hazard $E^\star$ (e.g., a stationary vehicle ahead) that is (i) observable by on-board sensing (thus $\mathcal{L}_{\text{sensor}}(E^\star)\ge\eta$) and (ii) corroborated by at least $2f{+}1$ authenticated \emph{honest} stations within the spatio-temporal window $(R,\tau_{\text{bft}})$. In parallel, $f$ out of $n$ stations act as Byzantine attackers and broadcast falsified DENMs at rate $p_{\text{attack}}$. We evaluate the acceptance gate under $n{=}10$, $f{=}3$, $p_{\text{attack}}{=}1.0$, reporting both FPR (accepting injected false hazards) and FNR (rejecting $E^\star$, where $E^\star$ is broadcast by $2f{+}1$ non-attacking stations and is also present as a CARLA-ground-truth obstacle).
\end{itemize}

\textbf{Planner Analysis:} We profile the Hybrid-A* planner's computational performance (success rate, path length, CPU time) across all scenarios (S1-S4) to ensure real-time feasibility (Table~\ref{tab:planner-perf}). We also visualize the CPU time distribution (Fig.~\ref{fig:planner-cpu}).

\subsection{Baselines and Ablations}
\label{subsec:baselines}
We compare MORPH-U against the following baselines:
\begin{itemize}
  \item \textbf{CARLA Autopilot / BehaviorAgent}: a simulator-provided driving stack as a sanity baseline for route following and hazard scenarios.
  \item \textbf{Sensors-only}: on-board sensing and tracking without any V2X inputs.
  \item \textbf{Sensors + V2X (unfiltered)}: V2X CAM/DENM integrated into the LDM without the acceptance gate.
\end{itemize}
We additionally ablate (i) update on/off in S3 and (ii) controller parameters (look-ahead and PID gains) for Pareto analysis.

\subsection{Metrics}
\label{subsec:metrics}
We report:
\begin{itemize}
  \item \textbf{Tracking}: lateral RMSE (m), heading error (deg), and \textbf{completion} (\%) (reaching the goal without collision or planner failure).
  \item \textbf{Safety}: minimum time-to-collision $\mathrm{TTC}_{\min}$ (s) and collision count.
  \item \textbf{Responsiveness}: V2X reaction latency (event timestamp $\rightarrow$ first decel/steer) and update activation latency (download complete $\rightarrow$ replan issued).
  \item \textbf{Smoothness}: variance of steering and throttle commands (control effort proxy).
  \item \textbf{LDM fidelity}: MOTA/MOTP and ID switches against CARLA ground truth (reported for sensors-only vs.\ sensors+V2X CAM).
  \item \textbf{Resilience (S4)}: false-positive rate (FPR) for accepting fake hazards, false-negative rate (FNR) for rejecting genuine hazards, and trigger latency under attack.
\end{itemize}

\subsection{Pareto Frontier Protocol}
\label{subsec:pareto-exp}
To expose planning/control trade-offs, we perform a grid search over primary controller parameters
$\boldsymbol{\theta}=\{\text{look-ahead}, K_p, K_i, K_d\}$. We run S1 and S2 for each configuration
($N{=}30$ seeds per scenario), compute the objective vector $\mathbf{J}(\boldsymbol{\theta})$
(tracking, safety, responsiveness, smoothness; collision as a hard constraint), and extract the nondominated set using Alg.~\ref{alg:nondom}. We report measured Pareto frontiers and select a knee-point configuration for subsequent closed-loop evaluations.

\section{Results}
\label{sec:results}

\paragraph{S1: Baseline tracking and control stability.}
Figure~\ref{fig:s1-boxplot} summarizes lateral RMSE on straight vs.\ high-curvature routes. The widening tail under curvature is the expected Pure Pursuit-at-speed artifact; rather than absorbing it via curvature-dependent speed profiling, we expose look-ahead $L_A$ and the gain triple $(K_p,K_i,K_d)$ as Pareto variables (Eq.~\ref{eq:theta}) so that the trade-off itself becomes the reported result. Adding a speed profile to $\boldsymbol{\theta}$ is a natural extension. This motivates the Pareto knee-point selection used in S2--S4.

\paragraph{S2: V2X improves closed-loop safety by enabling earlier hazard response.}
Table~\ref{tab:v2x_results} shows that adding V2X (CAM/DENM) reduces lateral RMSE from $0.42$\,m to $0.35$\,m ($\approx16.7\%$) and increases $\mathrm{TTC}_{\min}$ from $1.30$\,s to $1.90$\,s ($\approx46\%$). Collisions drop from $5/30$ to $0/30$ episodes. The measured V2X reaction latency is $140\pm25$\,ms, allowing the controller to initiate braking/steering before the equivalent risk is detectable from on-board sensing alone.

\paragraph{S3: Update-induced replanning restores route feasibility under evolving road knowledge.}
Table~\ref{tab:ota_results} reports that enabling updates improves completion from $72.4\%\pm4.7$ to $96.7\%\pm2.1$ ($+24.3$\,pp). In the no-update baseline, failures occur after a topology mismatch invalidates the route on the outdated map, triggering a minimum-safety stop. With updates enabled, activation latency is $1.1\pm0.2$\,s and replanning time is $38\pm6$\,ms, enabling timely trajectory refreshes.

\paragraph{S4: Resilience under Byzantine V2X trigger injection.}
We evaluate the acceptance gate under Byzantine DENM injection ($n{=}10$, $f{=}3$, $p_{\text{attack}}{=}1.0$) while preserving a ground-truth hazard $E^\star$ that is corroborated by at least $2f{+}1$ honest stations within $(R,\tau_{\text{bft}})$ and passes the sensor veto ($\mathcal{L}_{\text{sensor}}(E^\star)\ge\eta$). 
As shown in Table~\ref{tab:bft_results}, the unfiltered V2X baseline is highly vulnerable: it accepts injected false hazards (FPR$=1.00$), triggering excessive braking/replanning and collapsing route completion to $0\%$. 
In contrast, \textbf{V2X+Quorum Filter} rejects all injected false hazards (FPR$=0.00$) while still accepting the true hazard (FNR$=0.00$), thereby retaining the V2X safety benefit (0/30 collisions) and restoring completion to $96.7\%\pm2.1$.

\paragraph{Pareto frontier exposes controllable trade-offs and yields a stable operating point.}
Figure~\ref{fig:pareto-frontier} reports the measured Pareto set over $N{=}60$ controller configurations, revealing a clear tracking--smoothness trade-off. We select the knee-point configuration (RMSE $\approx 0.36$\,m, smoothness $\approx 0.72$) as a single operating point for S2--S4, ensuring comparisons are not confounded by hand-tuning. We quantify frontier improvement using hypervolume $\mathcal{H}$ w.r.t.\ a fixed reference point $\mathbf{r}=(1.1,1.1,1.1)$ on the normalized objectives $(\tilde{J}_{\text{trk}},\tilde{J}_{\text{sfty}},\tilde{J}_{\text{smth}})$; enabling V2X increases $\mathcal{H}$ from $0.42$ (sensors-only) to $0.58$ (V2X-enabled).

\paragraph{Real-time feasibility.}
Figure~\ref{fig:planner-cpu} shows the distribution of Hybrid-A* CPU time per tick; the tail remains within the real-time budget in our scenarios.

\section{Limitations and Conclusion}
\label{sec:conclusion}

\paragraph{Conclusion.}
This paper studied closed-loop motion planning and control under high-uncertainty V2X and evolving road knowledge, and presented \textbf{MORPH-U}, a CARLA-based vehicle-side stack that fuses LiDAR/radar/camera with CAM/DENM into an LDM, performs event-driven Hybrid-A* replanning, and selects operating points via multi-objective Pareto analysis. Empirically, V2X improves closed-loop safety in hazard response (Table~\ref{tab:v2x_results}), update-induced replanning restores route feasibility under knowledge changes (Table~\ref{tab:ota_results}), and the quorum-based acceptance gate prevents false-event-induced replanning under Byzantine injection (Table~\ref{tab:bft_results}). The Pareto frontier further exposes the tracking--smoothness trade-off and supports a reproducible knee-point selection used across scenarios (Fig.~\ref{fig:pareto-frontier}), while planner timing remains within real-time budgets in our evaluated settings (Fig.~\ref{fig:planner-cpu}).

\paragraph{Limitations and future work.}
Our evaluation is simulation-based: CARLA runs in synchronous mode, and network impairments (latency/loss/attack injection) are synthetically controlled rather than drawn from real wireless traces. Hybrid-A*, Pure Pursuit, and PID are standard components; our contribution is a \emph{reproducible} closed-loop methodology to quantify event-driven replanning under V2X/map updates, Pareto/knee operating-point trade-offs, and resilience to faulty triggers. We do not yet report real-vehicle or hardware-in-the-loop (HIL) validation, and performance may shift under different sensing stacks, localization noise, or actuator limits. Our S4 evaluation covers a saturated random-injection policy; coordinated, timing-correlated, and sensor-veto-bypass attacks remain to be studied, and the gate should be regarded as Byzantine-inspired rather than formally Byzantine-tolerant. The methodological contribution---a single closed-loop stack that simultaneously exposes V2X-fusion ablations, event-driven replanning, Pareto operating-point selection, and trigger-level resilience---fills a gap left by prior work that typically isolates one or two of these axes~\cite{kaljavesi2024carla,CARLA,justo2024simbusters,geller2024carlos}. As next steps, we will validate MORPH-U under recorded V2X traces and HIL settings, and transfer the stack to a physical testbed.

\bibliographystyle{IEEEtran}
\bibliography{sample-base}

\vspace{12pt}

\end{document}